%% file: main.tex
\documentclass[10pt,twocolumn,letterpaper]{article}
\usepackage[pagenumbers]{cvpr}
\input{preamble}
\definecolor{cvprblue}{rgb}{0.21,0.49,0.74}
\usepackage[breaklinks,colorlinks,allcolors=cvprblue]{hyperref}
\hypersetup{pdftitle={Conflict-Aware Multi-Teacher On-Policy Distillation in Aligned Action Spaces for Cross-Game Generalization},
  pdfauthor={Jing Yi, Cehao Yang, Hualei Zhang, Haitao Li}}
\title{Conflict-Aware Multi-Teacher On-Policy Distillation in Aligned Action Spaces\\for Cross-Game Generalization}
\author{Jing Yi\textsuperscript{1}\quad
Cehao Yang\textsuperscript{2}\quad
Hualei Zhang\textsuperscript{3}\quad
Haitao Li\textsuperscript{1}\\[.4em]
\textsuperscript{1}LIGHTSPEED\quad
\textsuperscript{2}HKUST(GZ)\quad
\textsuperscript{3}HKUST
}
\begin{document}
\setlength{\abovedisplayskip}{6.4pt plus 2pt minus 2pt}
\setlength{\belowdisplayskip}{6.4pt plus 2pt minus 2pt}
\setlength{\abovedisplayshortskip}{4.2pt plus 2pt minus 2pt}
\setlength{\belowdisplayshortskip}{4.2pt plus 2pt minus 2pt}
\makeatletter
\renewcommand{\paragraph}{\@startsection{paragraph}{4}{\z@}{1.76ex \@plus0.4ex \@minus.2ex}{-1em}{\normalfont\normalsize\bfseries}}
\makeatother
\maketitle
\input{sections/00_abstract}
\input{sections/01_introduction}
\input{sections/02_related_work}
\input{sections/03_method}
\input{sections/04_experiments}
\input{sections/05_discussion}
\input{sections/07_llm_usage}
{\small\bibliographystyle{ieeenat_fullname}\bibliography{references}}
\onecolumn
\appendix
\setcounter{figure}{0}
\setcounter{table}{0}
\renewcommand{\thefigure}{S\arabic{figure}}
\renewcommand{\thetable}{S\arabic{table}}
\renewcommand{\theHfigure}{appendix.\arabic{figure}}
\renewcommand{\theHtable}{appendix.\arabic{table}}
\input{sections/06_supplement}
\end{document}

%% file: preamble.tex
\usepackage[T1]{fontenc}
\usepackage[utf8]{inputenc}
\usepackage{mathtools}
\usepackage{array}
\usepackage{tabularx}
\usepackage{multirow}
\usepackage{colortbl}
\usepackage{microtype}

\newcommand{\method}{\textsc{CAMOPD}\xspace}
\newcommand{\sourcegames}{\mathcal{G}_{\mathrm{src}}}

\AtEndPreamble{
    \crefname{table}{Table}{Tables}
    \Crefname{table}{Table}{Tables}
    \crefname{figure}{Fig.}{Figs.}
    \Crefname{figure}{Fig.}{Figs.}
}

%% file: sections/00_abstract.tex
\begin{abstract}
Zero-shot cross-game generalization aims to consolidate accessible source-game experts into a single vision-language model (VLM) that controls unseen games from raw pixels without target-game parameter updates. While multi-teacher on-policy distillation (MOPD) mitigates offline exposure bias and single-game reinforcement learning (RL) overfitting, scaling it to autoregressive VLMs faces two bottlenecks. First, \emph{token-to-action misalignment} from subword length bias distorts discrete action probabilities and uncertainty margins. Second, \emph{multi-teacher policy conflict} arises as source teachers under incompatible dynamics issue contradictory closed-loop corrections, causing negative transfer. We propose \method (\textbf{C}onflict-\textbf{A}ware \textbf{M}ulti-Teacher \textbf{O}n-\textbf{P}olicy \textbf{D}istillation), a post-training framework that resolves multi-teacher policy conflict within an aligned semantic action space. \method first uses \emph{complete-action alignment} to normalize command completion likelihoods over each game's legal action set, forming a calibrated categorical policy that provides reliable uncertainty margins for selective on-policy retention and visual specialist probes that can be evaluated across source games. To resolve policy conflict without target-game data, \emph{conflict-aware multi-teacher consolidation} cross-evaluates these specialists across source games to capture both visual identifiability and inter-teacher compatibility, upweighting compatible teachers while suppressing conflicting corrections. On GameWorld's temporal and directional transfer suites with Qwen3.5-4B, \method achieves 26/30 (86.7\%) and 24/30 (80.0\%) Task-1 zero-shot success on unseen GeoDash and Doodle Jump, outperforming prompting, weight-merging, and multi-task baselines.
\end{abstract}

%% file: sections/01_introduction.tex
\section{Introduction}
\label{sec:introduction}

Vision-language models (VLMs) increasingly enable generalist agents to control
interactive environments from raw
pixels~\cite{reed2022gato,brohan2023rt2,kim2024openvla,deepmind2024sima,bytedance2025gametars,magne2026nitrogen},
with GameWorld~\cite{ouyang2026gameworld} standardizing state-verifiable
browser-game evaluation. Beyond in-distribution training
environments, open-ended deployment on user-generated content (UGC)
platforms, indie web portals, and HTML5 mini-games requires
\emph{cross-game generalization}: transferring reusable visual-motor skills to unseen games without target-game parameter updates. Because accessing private simulator state or running per-game RL for every new title is infeasible at deployment, we consolidate source-game teachers into a single pixel-only student operating solely on screenshots and public instructions.

\begin{figure*}[t]
    \centering
    \includegraphics[width=1\linewidth]{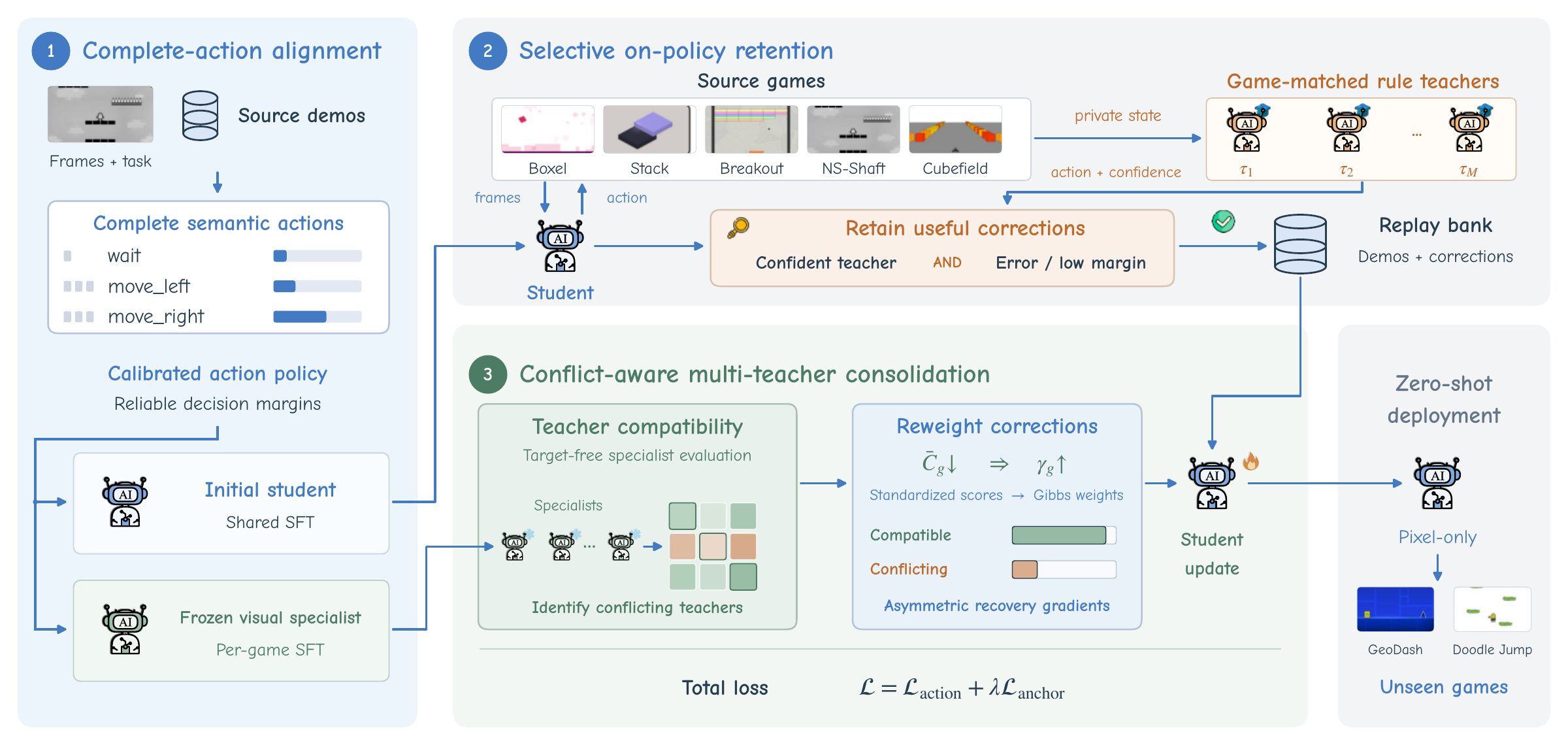}
    \vspace{-1.8em}
    \caption{\textbf{Overview of \method.}
\textbf{(1) Complete-action alignment} scores complete semantic actions to form a calibrated policy with reliable margins, initializing the shared student and frozen per-game specialists.
\textbf{(2) Selective on-policy retention} rolls out the student with rule teachers ($\tau_1, \dots, \tau_M$) to retain high-confidence teacher corrections on student-error or low-margin states in a replay bank.
\textbf{(3) Conflict-aware multi-teacher consolidation} cross-evaluates specialists without target data ($\bar{C}_g\!\downarrow \;\Rightarrow \gamma_g\!\uparrow$) to reweight asymmetric recovery gradients under $\mathcal{L} = \mathcal{L}_{\mathrm{action}} + \lambda\mathcal{L}_{\mathrm{anchor}}$.
\textbf{(4) Zero-shot deployment} runs the consolidated pixel-only student on unseen games.}
    \label{fig:framework}
    \vspace{-0.6em}
\end{figure*}

Adapting VLMs for zero-shot cross-game generalization faces challenges in both
action grounding and closed-loop training. At the decision interface,
autoregressive VLMs exhibit \textbf{token-to-action misalignment}: factorizing
commands into variable-length subword sequences over an open vocabulary biases
raw likelihoods toward shorter strings rather than optimal actions, distorting
action probabilities and uncertainty margins. Furthermore, conventional
post-training struggles under closed-loop transfer: offline supervised
fine-tuning (SFT) compounds errors once the agent drifts into unfamiliar
states~\cite{ross2011dagger}, whereas traditional on-policy
RL~\cite{schulman2017ppo,tong2026gamerl} either requires costly trial-and-error
in the target game or overfits to a single source game. Consolidating diverse
source experts via {multi-teacher on-policy distillation
(MOPD)}~\cite{ma2026mopd} thus offers a natural paradigm to learn closed-loop
recovery without target-game access.

However, scaling MOPD across heterogeneous source games exposes a critical
bottleneck: \textbf{multi-teacher policy conflict}. Source games often obey
incompatible control rules in visually similar situations---for example,
steering \emph{toward} foreground geometry to land on platforms (e.g.,
\emph{NS-Shaft}) versus steering \emph{away} from foreground blocks to dodge
obstacles (e.g., \emph{Cubefield}). Combining such contradictory teacher
corrections indiscriminately---via uniform MOPD~\cite{ma2026mopd}, gradient
projection~\cite{yu2020pcgrad}, or weight-space
merging~\cite{ilharco2023taskarithmetic,yadav2023ties,yu2024dare,deep2024della,jang2024modelstock,jin2023regmean,fu2026mergevla}---pulls
the shared student in opposite directions and causes severe negative transfer.
Identifying which teachers conflict without target-game data is also
non-trivial: unlike language-domain MOPD where every large language model (LLM) teacher can score any
text prompt~\cite{sun2026trustmopd}, privileged game teachers read
game-specific internal simulator variables (e.g., ball coordinates or tile
grids) rather than raw pixels, so a teacher for one game cannot be executed on
another game to measure inter-teacher agreement.

To address these challenges, we propose \textbf{\method}
(\textbf{C}onflict-\textbf{A}ware \textbf{M}ulti-Teacher \textbf{O}n-\textbf{P}olicy
\textbf{D}istillation), which resolves multi-teacher conflict within an aligned
semantic action space (\cref{fig:framework}). Rather than generating raw
subwords, \emph{Complete-Action Alignment} scores full legal commands per token
to establish a calibrated student policy and visual specialist proxies for the
privileged rule teachers. \emph{Selective On-Policy Retention} then rolls out
the student across source games, querying teachers only on uncertain or
erroneous boundary states~\cite{zhang2016safedagger}. Finally, because visual
specialists operate on pixels rather than private simulator variables,
\emph{Conflict-Aware Multi-Teacher Consolidation} cross-evaluates them across
peer source games to estimate teacher reliability and compatibility without
target-game data. At deployment, all teachers and probes are discarded, leaving
a single pixel-only Qwen3.5-4B policy evaluated zero-shot on GameWorld.
Our contributions are summarized as follows:
\begin{itemize}
    \item \textbf{Complete-action alignment.} We eliminate subword length bias via length-normalized candidate scoring over each game's legal action set, yielding calibrated decision margins for selective on-policy retention and enabling pixel-only specialist probes across distinct action sets.
    \item \textbf{Conflict-aware multi-teacher consolidation.} We formulate a target-free disagreement matrix by cross-evaluating source specialists, coupling in-game visual identifiability with cross-game control compatibility to reweight on-policy recovery gradients before consolidation.
    \item \textbf{Zero-shot transfer and boundary analysis.} \method reaches 86.7\% and 80.0\% Task-1 success on unseen GeoDash and Doodle Jump, surpassing standalone specialists, parameter mergers, and gradient surgery, while causal and cross-genre analyses characterize grounding and transfer boundaries.
\end{itemize}

%% file: sections/02_related_work.tex
\section{Related Work}
\label{sec:related}

\subsection{Generalist Game Agents and VLA Training}
Prior multimodal game and vision-language-action (VLA) control spans three agent families and distinct
benchmarks. (i) \emph{Large-scale trajectory pretraining}:
Gato~\cite{reed2022gato}, RT-2~\cite{brohan2023rt2}, and
OpenVLA~\cite{kim2024openvla} emit discretized control tokens, while
SIMA~\cite{deepmind2024sima}, Game-TARS~\cite{bytedance2025gametars}, and
NitroGen~\cite{magne2026nitrogen} scale behavior cloning (BC) or continual
pretraining from gameplay demonstrations (native keyboard--mouse trajectories or
flow-matched gamepad chunks). (ii) \emph{Training-free prompting}, such as
Cradle~\cite{tan2024cradle}, couples VLM reasoning with perception and
pause-and-plan modules. (iii) \emph{Online RL post-training}---RL4VLM~\cite{zhai2024rl4vlm},
VAGEN~\cite{wang2025vagen}, and Game-RL~\cite{tong2026gamerl}---fine-tunes VLMs
with environment rewards~\cite{schulman2017ppo}. For evaluation,
Orak~\cite{park2025orak} targets commercial games via Model Context Protocol
(MCP) textual states and agentic tools, whereas
GameWorld~\cite{ouyang2026gameworld} benchmarks pixel-only control in paused
browser sandboxes with state-verifiable progress. We adopt GameWorld's
pixel-only setting to study zero-shot consolidation of source teachers without
massive video pretraining or target-game RL.

\subsection{Multi-Teacher On-Policy Distillation}
Offline distillation compresses RL policies~\cite{rusu2015policy,parisotto2016actormimic}
or visual encoders (MoVE-KD~\cite{cao2025movekd}) on fixed data, whereas
interactive imitation mitigates rollout drift via online queries
(DAgger~\cite{ross2011dagger}), uncertainty gating
(SafeDAgger~\cite{zhang2016safedagger}), or privileged-to-visual
supervision~\cite{chen2020learningbycheating}. In autoregressive models,
on-policy distillation (OPD) similarly supervises student rollouts across
language generation (MiniLLM~\cite{gu2024minillm}, GKD~\cite{agarwal2024gkd}),
VLM reasoning (VOLD~\cite{bousselham2025vold}), and VLA control
(VLA-OPD~\cite{zhong2026vlaopd}). Multi-teacher variants aggregate domain
experts (MOPD~\cite{ma2026mopd}, MAD-OPD~\cite{wang2026madopd}) or mitigate
conflict via token reliability (TrustMOPD~\cite{sun2026trustmopd}), subspace
constraints (PMOPD~\cite{liu2026pmopd}), and update projection
(UP-MOPD~\cite{zhu2026upmopd}). Yet these methods assume homogeneous
teachers over a shared vocabulary, whereas \method resolves conflict across
heterogeneous game teachers via source-only visual probing.

\subsection{Action Grounding, Merging, and Routing}
Executable grounding constrains outputs via grammar
decoding~\cite{scholak2021picard}, candidate scoring~\cite{yao2020calm},
affordances~\cite{ahn2022saycan}, or uncertainty
calibration~\cite{ren2023knowno}; we length-normalize candidate likelihoods to
remove multi-token bias. To combine Low-Rank Adaptation (LoRA)~\cite{hu2022lora}
adapters without target updates, static mergers---RegMean~\cite{jin2023regmean}, Task
Arithmetic~\cite{ilharco2023taskarithmetic}, TIES~\cite{yadav2023ties},
DARE~\cite{yu2024dare}, DELLA~\cite{deep2024della}, Model
Stock~\cite{jang2024modelstock}, and AdaMerging~\cite{yang2024adamerging}---fuse
weights, and TIME~\cite{dziadzio2025time} merges sequential multimodal tasks.
Alternatively, PCGrad~\cite{yu2020pcgrad} and CAGrad~\cite{liu2021cagrad} project
conflicting gradients, LoraHub~\cite{huang2024lorahub} and
Twin-Merging~\cite{lu2024twinmerging} compose LoRA modules at test time,
MergeVLA~\cite{fu2026mergevla} pairs task-masked merging with a test-time
router, and HiMoE-VLA~\cite{du2025himoe} co-trains hierarchical action-space
Mixture-of-Experts (MoEs). However, these methods treat conflicting experts
equally or rely on test-time routing, whereas \method asymmetrically
consolidates teachers in action space into one student.

%% file: sections/03_method.tex
\section{Method}
\label{sec:method}

\method consolidates heterogeneous source teachers into a single pixel-only VLM
through three progressive components (\cref{fig:framework}; Algorithm~1 in
Appendix~\ref{sec:supp-alg}). Following the problem setup (\cref{sec:problem}), we
detail \emph{Complete-Action Alignment} (\cref{sec:action-alignment}),
\emph{Selective On-Policy Retention} (\cref{sec:mopd}), and
\emph{Conflict-Aware Multi-Teacher Consolidation}
(\cref{sec:conflict-consolidation}).

\subsection{Problem Formulation}
\label{sec:problem}

\paragraph{Source and target POMDPs.}
Let $\sourcegames = \{g_1, \dots, g_M\}$ denote the set of $M$ source training
games and let $g^\star \notin \sourcegames$ denote an unseen held-out target
game. Each game $g \in \sourcegames \cup \{g^\star\}$ is formulated as a
partially observable Markov decision process (POMDP) with simulator state
$s_t \in \mathcal{S}_g$, raw RGB screenshot observation $o_t \in \mathcal{O}_g$,
and finite legal semantic action set $\mathcal{A}_g$. The deployed student does
not access private simulator state $s_t$; at decision step $t$, it receives the
multimodal context $x_t$ consisting of the last $h \in \{1, 2\}$ screenshots
$o_{t-h+1:t} = (o_{t-h+1}, \dots, o_t)$ and the game's public task instruction
prompt $q_g$ specifying $\mathcal{A}_g$:
\begin{equation}
    x_t = (o_{t-h+1:t}, q_g).
    \label{eq:context}
\end{equation}
Frame stacks ($h=2$) provide velocity cues for steering, while single frames
($h=1$) suffice for timing.

\paragraph{Privileged source teachers and held-out targets.}
During source collection ($g \in \sourcegames$), each source game is paired
with a programmatic rule teacher
$\tau_g : \mathcal{S}_g \to \mathcal{A}_g \times (0, 1]$ that reads
privileged state $s_t \in \mathcal{S}_g$ and outputs reference action
$a_t^\star \in \mathcal{A}_g$ and criticality confidence
$c_t \in (0, 1]$ (Appendix~\ref{sec:supp-prompts-teachers}). Rolling out $\tau_g$
yields the initial SFT dataset
$\mathcal{D}_0 = \bigcup_{g \in \sourcegames} \mathcal{D}_g$ with
$\mathcal{D}_g = \{ (x_t, \mathcal{A}_g, a_t^\star, c_t) \}_{t=1}^{N_g}$, where
training splits are used for policy updates and held-out action-balanced
validation splits $\mathcal{D}_g^{\mathrm{val}}$ are reserved for cross-source
conflict probing. Target transitions, screenshots, and rewards are excluded from
policy updates, and teacher allocation weights are computed from source
validation data alone.

\begin{table*}[!t]
\centering
\caption{\textbf{Transfer suites and dataset statistics.}
Each suite consolidates heterogeneous source teachers ($\sourcegames$) into a
single pixel-only student evaluated zero-shot on an unseen target game
($g^\star$). Training samples report the initial SFT split ($\mathcal{D}_0$)
and retained on-policy corrections ($\mathcal{Q}$).}
\label{tab:suite-roles}
\setlength{\tabcolsep}{0pt}
\renewcommand{\arraystretch}{1.02}
\footnotesize
\begin{tabular*}{\textwidth}{@{\extracolsep{\fill}}llllll@{}}
\toprule
Suite & Control Axis & Source Teachers ($\sourcegames$) & Samples ($\mathcal{D}_0$ / $\mathcal{Q}$) & Held-Out Target ($g^\star$) & Target Tasks (Threshold) \\
\midrule
A & Temporal (when to act) & Boxel Rebound, Stack & 3,519 / 392 & GeoDash (\texttt{wait}, \texttt{jump}) & T1 ($\ge 50$), T2 ($\ge 100$) \\
B & Directional (where to steer) & Breakout, NS-Shaft, Cubefield & 2,104 / 107 & Doodle Jump (\texttt{wait}, \texttt{move\_left/right}) & T1 ($\ge 200$), T2 ($\ge 500$) \\
\bottomrule
\end{tabular*}
\vspace{-0.4em}
\end{table*}

\subsection{Complete-Action Alignment}
\label{sec:action-alignment}

\paragraph{Length-normalized action scoring.}
Because each game exposes a distinct legal action set $\mathcal{A}_g$, we align
decisions in semantic action space rather than open-vocabulary token space.
Standard autoregressive behavior cloning factorizes each command string into a
variable-length subword sequence $y(a) = (y_1, \dots, y_{L_a})$ of length $L_a$
and optimizes token-level cross-entropy over an open language vocabulary, which
tends to bias raw generation likelihoods toward shorter or lexically favored
strings rather than optimal control actions.
To align scoring with atomic executable decisions, we attach a shared LoRA
adapter~\cite{hu2022lora} with parameters $\theta$ to the VLM attention and multilayer perceptron (MLP)
projections and evaluate each legal command $a \in \mathcal{A}_g$ as a complete
semantic unit. Given multimodal context $x$, we render the prompt under a
deterministic non-thinking template (Appendix~\ref{sec:supp-prompts-teachers})
and construct $|\mathcal{A}_g|$ candidate sequences by appending each structured JSON
command completion $y(a)$. Masking prompt tokens, we compute the
length-normalized completion score over all $L_a$ tokens:
\begin{equation}
    s_\theta(a \mid x) = \frac{1}{L_a}
    \sum_{\ell=1}^{L_a} \log p_\theta\!\left(y_\ell(a) \;\middle|\; x, y_{<\ell}(a)\right),
    \label{eq:score}
\end{equation}
where $p_\theta(y_\ell(a) \mid x, y_{<\ell}(a))$ is the autoregressive
completion probability. Masking prompt tokens avoids context dilution, while
dividing by $L_a$ scores short (e.g., \texttt{wait}) and multi-token
(e.g., \texttt{move\_right}) commands per token.

\paragraph{Categorical policy and model initialization.}
Applying a softmax with temperature $T > 0$ over
$\mathcal{A}_g$ maps the command scores into a differentiable categorical policy
$\pi_\theta^{(T)}$ ($\pi_\theta$ at $T = 1$):
\begin{equation}
    \pi_\theta^{(T)}(a \mid x, \mathcal{A}_g) =
    \frac{\exp\!\left(s_\theta(a \mid x)/T\right)}
    {\sum_{a' \in \mathcal{A}_g} \exp\!\left(s_\theta(a' \mid x)/T\right)}.
    \label{eq:policy}
\end{equation}
Offline action-aligned SFT initializes the shared student adapter
$\theta_{\mathrm{SFT}}$ on $\mathcal{D}_0$ via complete-action negative
log-likelihood (NLL):
\begin{equation}
    \mathcal{L}_{\mathrm{SFT}}(\theta) =
    \mathbb{E}_{(x, \mathcal{A}_g, a^\star) \sim \mathcal{D}_0}
    \left[ -\log \pi_\theta(a^\star \mid x, \mathcal{A}_g) \right].
    \label{eq:sft}
\end{equation}
In parallel, minimizing \cref{eq:sft} on each single-game dataset
$\mathcal{D}_g$ yields a frozen visual specialist probe
$\pi_{T_g}(a \mid x, \mathcal{A}_g)$ that acts as a pixel-based proxy for
$\tau_g$ and can be evaluated on any peer game $g'$ over $\mathcal{A}_{g'}$.

\paragraph{Action-simplex gradients and margin calibration.}
Differentiating \cref{eq:sft} with respect to each candidate score $s_\theta(a
\mid x)$ ($a \in \mathcal{A}_g$, $T=1$) yields
\begin{equation}
    \frac{\partial \!\left(-\log \pi_\theta(a^\star \mid x, \mathcal{A}_g)\right)}
    {\partial s_\theta(a \mid x)}
    = \pi_\theta(a \mid x, \mathcal{A}_g) - \mathbb{I}[a = a^\star].
    \label{eq:action-grad}
\end{equation}
Standard token-level cross-entropy normalizes over the entire open vocabulary
at every subword step, diluting gradients across non-executable tokens and
biasing likelihoods by length $L_a$. In contrast, \cref{eq:action-grad}
restricts competition to the legal action set $\mathcal{A}_g$: raising
$s_\theta(a^\star \mid x)$ directly pushes down each competing command $a \ne
a^\star$ in proportion to its probability $\pi_\theta(a \mid x, \mathcal{A}_g)$.
Dividing by $L_a$ also calibrates top-two score margins on a per-token scale,
providing a length-unbiased confidence signal to identify low-margin boundary
states for selective on-policy retention (\cref{sec:mopd}). Moreover, unlike
fixed-dimension action heads, scoring command strings via the shared VLM head
constructs $\pi_\theta(\cdot \mid x, \mathcal{A}_{g^\star})$ for any unseen
target game $g^\star$ without target-head retraining.

\subsection{Selective On-Policy Retention}
\label{sec:mopd}

Offline demonstrations $\mathcal{D}_0$ cover only teacher-visited trajectories,
leaving the student susceptible to compounding drift on unfamiliar test-time
states~\cite{ross2011dagger}. Within MOPD, we roll out the shared student
$\pi_\theta$ across source games $g \in \sourcegames$. At each visited state
$x_t$, the student executes
$\hat{a}_t = \arg\max_{a \in \mathcal{A}_g} \pi_\theta(a \mid x_t, \mathcal{A}_g)$
with calibrated margin
$m_\theta(x_t) = s_\theta(\hat{a}_t \mid x_t) - \max_{a \in \mathcal{A}_g \setminus \{\hat{a}_t\}} s_\theta(a \mid x_t) \ge 0$,
while the game-matched rule teacher $\tau_g(s_t)$ returns reference action
$a_t^\star$ and confidence $c_t$. Rather than retaining all rollout frames as
in standard DAgger~\cite{ross2011dagger}, a transition enters the on-policy
correction buffer $\mathcal{Q}_g$~\cite{zhang2016safedagger} when the teacher
is confident ($c_t \ge c_{\min} > 0$) and the student either errs ($\hat{a}_t
\ne a_t^\star$) or exhibits a low margin ($m_\theta(x_t) < \delta$):
\begin{equation}
    \mathcal{I}_{\mathrm{retain}}(x_t, s_t) =
    \mathbb{I}[c_t \ge c_{\min}] \cdot
    \mathbb{I}\!\left[\hat{a}_t \ne a_t^\star \;\lor\; m_\theta(x_t) < \delta\right].
\label{eq:query}
\end{equation}
Here $c_t \ge c_{\min}$ filters out incomplete or degenerate simulator states,
while $\hat{a}_t \ne a_t^\star \lor m_\theta(x_t) < \delta$ omits
routine agreement frames (80\%--90\% of rollout steps) to retain student-induced
errors and near-boundary states. Because the retained set $\mathcal{Q} = \bigcup_{g \in \sourcegames}
\mathcal{Q}_g$ accounts for at most $10\%$ of the combined data
(\cref{tab:suite-roles}), training on $\mathcal{Q}$ alone risks forgetting
nominal control, so we consolidate on $\mathcal{D}_0 \cup \mathcal{Q}$.

\subsection{Conflict-Aware Multi-Teacher Consolidation}
\label{sec:conflict-consolidation}

\paragraph{Multi-teacher gradient conflict in uniform MOPD.}
When training on $\mathcal{D}_0 \cup \mathcal{Q}$, we amplify retained
corrections $i \in \mathcal{Q}$ by an on-policy weight $w_q \ge 1$ so this small
boundary set is not overwhelmed by $\mathcal{D}_0$. However, uniform
MOPD~\cite{ma2026mopd} treats all source teachers equally by applying the
\emph{same} $w_q$ across every $g \in \sourcegames$: it updates $\theta$ along
$\bar{v}_0(\theta) + w_q \sum_{g \in \sourcegames} \bar{v}_g(\theta)$, combining
the aggregated action-loss gradient $\bar{v}_0(\theta)$ on offline
demonstrations $\mathcal{D}_0$ with each teacher's on-policy recovery gradient
$\bar{v}_g(\theta)$ on $\mathcal{Q}_g$. By the
chain rule on \cref{eq:action-grad}, when an incompatible teacher $g^-$ and a
compatible teacher $g^+$ prescribe opposing actions in visually similar states
(e.g., steering \emph{toward} platforms in \emph{NS-Shaft} vs.\ \emph{away} from
blocks in \emph{Cubefield}), they drive action scores in opposite directions and
induce conflicting gradients ($\langle \bar{v}_{g^-}(\theta), \bar{v}_{g^+}(\theta)
\rangle < 0$). Uniform pooling or symmetric gradient surgery
(PCGrad~\cite{yu2020pcgrad}) gives incompatible and transferable teachers equal
priority, attenuating $\bar{v}_{g^+}(\theta)$. Resolving this conflict requires
reweighting teachers asymmetrically before gradient aggregation, yet estimating
inter-teacher compatibility without target data is non-trivial because rule
teacher $\tau_g(s_t)$ depends on private simulator state and cannot run on peer
games $g' \ne g$.

\paragraph{Cross-source specialist disagreement probing.}
To bypass private simulator states, we use the frozen visual specialist probes
$\{\pi_{T_g}\}$ trained in \cref{sec:action-alignment} as pixel-based proxies
for $\{\tau_g\}$. Because each $\pi_{T_g}$ is a LoRA adapter over the shared VLM
backbone that scores candidate commands directly from screenshots, it can be evaluated on any peer game's
action-balanced validation slice $\mathcal{D}_{g'}^{\mathrm{val}}$ ($g' \in
\sourcegames$) over native command strings $a \in \mathcal{A}_{g'}$ under prompt
$q_{g'}$. Without requiring identical action names or hand-crafted aliases,
$1/L_a$ normalization removes multi-token bias on $\mathcal{A}_{g'}$ (restoring
a calibrated base-VLM prior) while using
pretrained verb semantics. Cross-evaluating all specialists yields the $M \times
M$ matrix $C_{g, g'}$ and row mean $\bar{C}_g = \frac{1}{M} \sum_{g' \in
\sourcegames} C_{g, g'}$:
\begin{equation}
    C_{g, g'} = \mathbb{E}_{(x, \mathcal{A}_{g'}, a^\star) \sim \mathcal{D}_{g'}^{\mathrm{val}}}
    \!\left[ -\log \pi_{T_g}(a^\star \mid x, \mathcal{A}_{g'}) \right]\!.
\label{eq:cross-source-nll}
\end{equation}
Decomposing $\bar{C}_g = \frac{1}{M}
C_{g,g} + \frac{1}{M}\sum_{g' \ne g} C_{g,g'}$ explains why source-only probing
predicts transfer even at $M{=}2$: it jointly penalizes (i) \emph{poor visual
identifiability} ($C_{g,g}$) when state rules are ill-posed from local pixels, and (ii) \emph{cross-game control
conflict} ($C_{g,g'}$, $g' \ne g$) from opposing dynamics.

\begin{table*}[!t]
\centering
\caption{\textbf{Zero-shot cross-game transfer on held-out target games (30 executions per task).}
We report task successes out of 30 (SR), T1--T2 Mean SR (\%) and PG (\%), and total failure resets under the Paused 100-Step Reset-on-Fail protocol. Best results (lowest resets among task-solving policies, $\mathrm{SR}>0$) are in \textbf{bold}.}
\label{tab:main-summary}
\setlength{\tabcolsep}{3.90pt}
\renewcommand{\arraystretch}{0.88}
\footnotesize
\begin{tabular}{@{}llcccccccccccccc@{}}
\toprule
& & \multicolumn{7}{c}{\textbf{Suite A: Held-Out GeoDash}} & \multicolumn{7}{c}{\textbf{Suite B: Held-Out Doodle Jump}} \\
\cmidrule(lr){3-9}\cmidrule(lr){10-16}
Method & Family & \multicolumn{2}{c}{Task 1} & \multicolumn{2}{c}{Task 2} & \multicolumn{2}{c}{\textbf{Mean}} & Resets$\downarrow$ & \multicolumn{2}{c}{Task 1} & \multicolumn{2}{c}{Task 2} & \multicolumn{2}{c}{\textbf{Mean}} & Resets$\downarrow$ \\
& & SR & PG & SR & PG & SR & PG & & SR & PG & SR & PG & SR & PG & \\
\midrule
Base VLM & pretrained & 0/30 & 92.0 & 0/30 & 46.0 & 0.0 & 69.0 & 66 & 3/30 & 63.3 & 0/30 & 17.3 & 5.0 & 40.3 & 69 \\
Pause-and-Plan CoT~\cite{tan2024cradle} & prompted VLM & 13/30 & 88.6 & 0/30 & 59.0 & 21.7 & 73.8 & 56 & 2/30 & 61.1 & 0/30 & 24.3 & 3.3 & 42.7 & 48 \\
Few-shot ICL & prompted VLM & 5/30 & 73.5 & 0/30 & 47.5 & 8.3 & 60.5 & 205 & 12/30 & 73.0 & 0/30 & 23.1 & 20.0 & 48.1 & 46 \\
\midrule
Boxel Rebound specialist & single teacher (A) & 14/30 & 84.2 & 0/30 & 49.7 & 23.3 & 67.0 & 69 & -- & -- & -- & -- & -- & -- & -- \\
Stack specialist & single teacher (A) & 0/30 & 30.0 & 0/30 & 15.0 & 0.0 & 22.5 & 3102 & -- & -- & -- & -- & -- & -- & -- \\
Breakout specialist & single teacher (B) & -- & -- & -- & -- & -- & -- & -- & 0/30 & 10.2 & 0/30 & 3.9 & 0.0 & 7.1 & 0 \\
Cubefield specialist & single teacher (B) & -- & -- & -- & -- & -- & -- & -- & 0/30 & 0.0 & 0/30 & 0.0 & 0.0 & 0.0 & 0 \\
NS-Shaft specialist & single teacher (B) & -- & -- & -- & -- & -- & -- & -- & 6/30 & 67.8 & 0/30 & 20.3 & 10.0 & 44.1 & 31 \\
\midrule
Task Arithmetic~\cite{ilharco2023taskarithmetic} & weight merge & 5/30 & 73.1 & 0/30 & 48.7 & 8.3 & 60.9 & 80 & 0/30 & 0.0 & 0/30 & 0.0 & 0.0 & 0.0 & 0 \\
TIES-Merging~\cite{yadav2023ties} & weight merge & 19/30 & 88.2 & 0/30 & 46.0 & 31.7 & 67.1 & 59 & 0/30 & 0.0 & 0/30 & 0.0 & 0.0 & 0.0 & 0 \\
DARE-TIES~\cite{yu2024dare} & weight merge & 13/30 & 83.5 & 0/30 & 58.8 & 21.7 & 71.1 & 51 & 0/30 & 1.0 & 0/30 & 0.0 & 0.0 & 0.5 & 0 \\
DELLA~\cite{deep2024della} & weight merge & 17/30 & 87.1 & 0/30 & \textbf{59.6} & 28.3 & 73.3 & 44 & 0/30 & 0.8 & 0/30 & 0.0 & 0.0 & 0.4 & 0 \\
Model Stock~\cite{jang2024modelstock} & weight merge & 3/30 & 92.8 & 0/30 & 48.8 & 5.0 & 70.8 & 60 & 2/30 & 49.9 & 0/30 & 18.6 & 3.3 & 34.2 & 66 \\
RegMean (diagonal)~\cite{jin2023regmean} & weight merge & 11/30 & 81.5 & 0/30 & 52.2 & 18.3 & 66.8 & 68 & 0/30 & 0.0 & 0/30 & 0.0 & 0.0 & 0.0 & 0 \\
Multi-Task PCGrad~\cite{yu2020pcgrad} & gradient surgery & 0/30 & 30.0 & 0/30 & 15.0 & 0.0 & 22.5 & 3002 & 0/30 & 16.4 & 0/30 & 7.6 & 0.0 & 12.0 & 125 \\
Feature-Routing MoE~\cite{lu2024twinmerging} & routed experts & 17/30 & 85.9 & 0/30 & 50.2 & 28.3 & 68.1 & 60 & 4/30 & 62.3 & 0/30 & 19.9 & 6.7 & 41.1 & 24 \\
\midrule
\rowcolor{cvprblue!10}
\textbf{\method (Ours)} & on-policy distillation & \textbf{26/30} & \textbf{95.5} & 0/30 & 58.0 & \textbf{43.3} & \textbf{76.7} & \textbf{36} & \textbf{24/30} & \textbf{91.9} & 0/30 & \textbf{39.4} & \textbf{40.0} & \textbf{65.7} & \textbf{12} \\
\bottomrule
\end{tabular}
\vspace{-0.5em}
\end{table*}

\begin{figure*}[!t]
    \centering
    \includegraphics[width=0.94\linewidth]{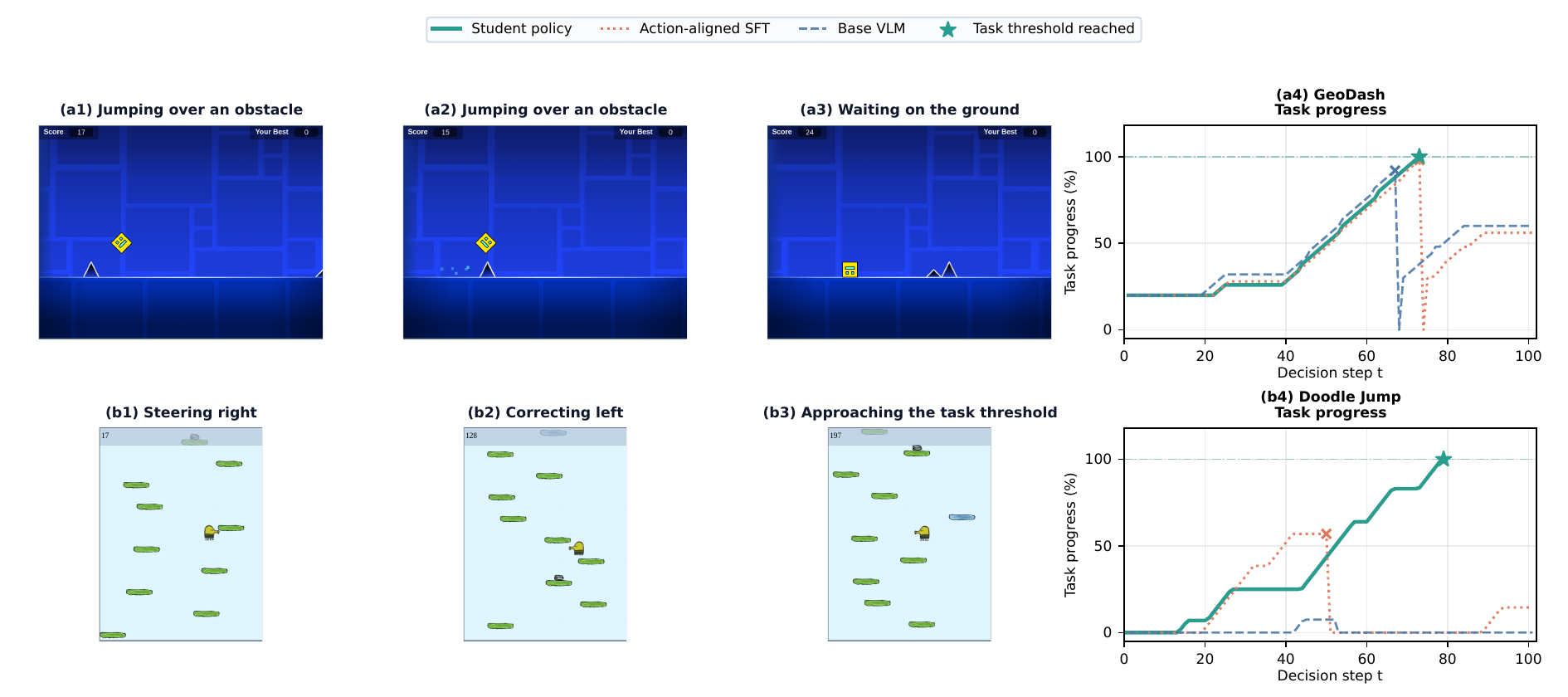}
    \vspace{-0.6em}
    \caption{\textbf{Qualitative closed-loop rollouts and Task-1 progress curves on unseen target games.}
    \textbf{Left:} Key frames on (a1--a3) \emph{GeoDash} and (b1--b3) \emph{Doodle Jump}.
    \textbf{Right:} Normalized task progress over the 100-decision horizon (a4, b4); sharp vertical drops to $0\%$ denote failure resets, whereas \method (\emph{Student policy}) reaches the $100\%$ threshold ($\star$).}
    \label{fig:results}
    \vspace{-0.6em}
\end{figure*}

\paragraph{Conflict-aware reweighting and anchored objective.}
To resolve policy conflict without target-game data, we map standardized
disagreement scores $(\bar{C}_g - \mu_C)/\sigma_C$ (with mean $\mu_C$, population std
$\sigma_C$, sharpness $\beta \ge 0$, and uniform prior floor $\epsilon \in [0,
1]$) into an $\epsilon$-regularized Gibbs distribution $\rho_g$ and mean-one
teacher multipliers $\gamma_g \in [\epsilon, M(1-\epsilon)+\epsilon]$:
\begin{equation}
\begin{split}
    z_g &= -\beta \frac{\bar{C}_g - \mu_C}{\sigma_C}, \qquad
    \gamma_g = M\,\rho_g, \\[-1pt]
    \rho_g &= (1 - \epsilon)\frac{\exp(z_g)}{\sum_{g' \in \sourcegames} \exp(z_{g'})} + \frac{\epsilon}{M}.
\end{split}
\label{eq:compatibility-gate}
\end{equation}
Standardizing $\bar{C}_g$ into $z_g$ makes $\beta$ invariant to action-set size
$|\mathcal{A}_g|$ and entropy scale across suites, while $\epsilon > 0$ bounds
$\gamma_g \ge \epsilon$ to preserve multi-source visual coverage.
With on-policy indicator $q_i = \mathbb{I}[i \in \mathcal{Q}]$ and floored
confidence $\tilde{c}_i = \max(c_{\mathrm{floor}}, c_i)$ ($c_{\mathrm{floor}} >
0$ avoiding zero sample weights), scaling $w_q$ by
$\gamma_{g_i}$ assigns each transition conflict-aware priority
$u_i^{\mathrm{ca}} = (1 + (w_q\gamma_{g_i} - 1)q_i)\tilde{c}_i$ and normalized
replay probability $P_{\mathrm{global}}(i) = u_i^{\mathrm{ca}} / Z_\gamma$
(where $Z_\gamma = \sum_{j \in \mathcal{D}_0 \cup \mathcal{Q}}
u_j^{\mathrm{ca}}$), yielding the asymmetric action-loss gradient:
\begin{equation}
    \nabla_\theta \mathcal{L}_{\mathrm{action}}(\theta)
    = \frac{1}{Z_\gamma}\,\bigg(
        \bar{v}_0(\theta)
        + w_q \sum_{g \in \sourcegames} \gamma_g\,\bar{v}_g(\theta)
    \bigg),
\label{eq:grad-decomp}
\end{equation}
where $\bar{v}_0(\theta)$ and $\bar{v}_g(\theta)$ aggregate the
$\tilde{c}_i$-weighted action NLL gradients over $\mathcal{D}_0$ and
$\mathcal{Q}_g$.
Because $\frac{1}{M}\sum_g \gamma_g = 1$ with $\gamma_{g^+} > 1 > \gamma_{g^-}$,
\cref{eq:grad-decomp} upweights compatible corrections and suppresses
conflicting ones ($\beta=0$ recovers uniform MOPD). The multiplier $\gamma_g$
scales only the on-policy term: consolidation starts from $\theta_{\mathrm{SFT}}$
under $\mathcal{L}_{\mathrm{anchor}}$, so $\bar{v}_0(\theta)$ on $\mathcal{D}_0$
is near saturation and maintains the shared visual representation, whereas
$w_q\sum_g\gamma_g\bar{v}_g(\theta)$ sets boundary recovery without discarding
offline samples. Finally, to absorb reweighted corrections while limiting policy
drift from $\theta_{\mathrm{ref}} = \theta_{\mathrm{SFT}}$, we update $\theta$
on minibatches $i \sim P_{\mathrm{global}}$ via the anchored loss ($\lambda \ge
0$, $T_{\mathrm{ref}} > 0$):
\begin{equation}
\begin{split}
    \mathcal{L}(\theta) &= \mathcal{L}_{\mathrm{action}}(\theta)
    + \lambda\,\mathcal{L}_{\mathrm{anchor}}(\theta), \\
    \mathcal{L}_{\mathrm{action}}(\theta) &= \mathbb{E}_{i \sim P_{\mathrm{global}}}
    \!\left[ -\log \pi_\theta(a_i^\star \mid x_i, \mathcal{A}_{g_i}) \right]\!, \\
    \mathcal{L}_{\mathrm{anchor}}(\theta) &= \mathbb{E}_{i \sim P_{\mathrm{global}}}
    \!\left[ T_{\mathrm{ref}}^2\,\mathrm{KL}\!\left(
        \pi_{\theta_{\mathrm{ref}}}^{(T_{\mathrm{ref}})}(x_i)
        \,\middle\|\,
        \pi_\theta^{(T_{\mathrm{ref}})}(x_i)
    \right) \right]\!,
\end{split}
\label{eq:amopd}
\end{equation}
where $\pi^{(T_{\mathrm{ref}})}(x_i)$ abbreviates
$\pi^{(T_{\mathrm{ref}})}(\cdot \mid x_i, \mathcal{A}_{g_i})$,
$\mathrm{KL}(\cdot\|\cdot)$ is the Kullback--Leibler divergence, and
$T_{\mathrm{ref}}^2$ preserves gradient scale. Matching the full reference
distribution via $\mathcal{L}_{\mathrm{anchor}}$ prevents entropy collapse under
amplified boundary updates ($w_q\gamma_g$). At deployment, all teachers and
probes are discarded, leaving one consolidated student $\pi_\theta$.

%% file: sections/04_experiments.tex
\section{Experiments}
\label{sec:experiments}

We evaluate \method on unseen target games under GameWorld's {Paused
100-Step Reset-on-Fail protocol}~\cite{ouyang2026gameworld}.

\subsection{Experimental Setup and Baselines}
\label{sec:setup}

\paragraph{Environments and evaluation protocol.}
As summarized in \cref{tab:suite-roles}, \textbf{Suite~A} evaluates temporal
intervention transfer from \emph{Boxel Rebound} and \emph{Stack} to held-out
\emph{GeoDash} ($\mathcal{A}_{g^\star} = \{\texttt{wait}, \texttt{jump}\}$,
$h{=}1$), while \textbf{Suite~B} evaluates directional steering transfer from
\emph{Breakout}, \emph{NS-Shaft}, and \emph{Cubefield} to held-out \emph{Doodle
Jump} ($\mathcal{A}_{g^\star} = \{\texttt{wait}, \texttt{move\_left}, \texttt{move\_right}\}$,
$h{=}2$). Each episode launches a fresh browser instance, pauses the simulator
during VLM inference, and executes up to 100 closed-loop decisions, resetting
upon terminal failure. We evaluate Tasks~1 and~2 over 30 executions each and
report success rate (\textbf{SR}), peak progress (\textbf{PG}, \%), T1--T2
mean, and \textbf{Resets}$\downarrow$.

\paragraph{Implementation details.}
All methods share Qwen3.5-4B with rank-16 LoRA~\cite{hu2022lora} on attention
and MLP projections (sequence length 2048, 65,536--401,408 pixels), optimized
using AdamW ($T = T_{\mathrm{ref}} = 1$). Selective retention uses
$c_{\min} = 0.75$ and $\delta = 0.25$, while conflict-aware consolidation uses
$\beta = 1$, $\epsilon = 0.1$, $c_{\mathrm{floor}} = 0.05$, and $w_q = 3$; we
set the anchor weight and learning rate to $\lambda = 0.5$ and $10^{-5}$ for
Suite~A, and to $\lambda = 1.0$ and $2\times 10^{-5}$ for Suite~B.

\paragraph{Baselines.}
We compare \method against four baseline families
(\cref{tab:main-summary}): (i) \emph{Prompted VLMs}: Base VLM,
Pause-and-Plan chain-of-thought (CoT)~\cite{tan2024cradle}, and few-shot
in-context learning (ICL);
(ii) \emph{Single-teacher specialists} trained on each source game;
(iii) \emph{Parameter-space mergers}: Task
Arithmetic~\cite{ilharco2023taskarithmetic}, TIES-Merging~\cite{yadav2023ties},
DARE-TIES~\cite{yu2024dare}, DELLA~\cite{deep2024della}, Model
Stock~\cite{jang2024modelstock}, and RegMean (diagonal)~\cite{jin2023regmean}; and
(iv) \emph{Multi-teacher optimization and routing}:
Multi-Task PCGrad~\cite{yu2020pcgrad} and Feature-Routing MoE~\cite{lu2024twinmerging}.

\subsection{Main Cross-Game Transfer Results}
\label{sec:results}

\Cref{tab:main-summary,fig:results} summarize zero-shot transfer performance
and representative closed-loop rollouts across both suites. Because reaching
the Task-1 target requires ${\sim}75$--$80$ steps (\cref{fig:results}(a4, b4)),
the doubled Task-2 thresholds (\cref{tab:suite-roles}) exceed the physical
reach of the 100-step horizon, yielding 0/30 Task-2 SR across all methods and
differentiating long-horizon control through Task-2 PG.

\begin{table*}[t]
\centering
\caption{\textbf{Component ablation across action alignment and conflict-aware consolidation (30 executions/task).}
Starting from \textbf{Token SFT}, we add \textbf{Complete-Action Alignment}, \textbf{Selective MOPD (Uniform)}, and \textbf{Conflict-Aware Consolidation} (\textbf{Full \method}).}
\label{tab:component-ablation}
\setlength{\tabcolsep}{3.28pt}
\renewcommand{\arraystretch}{0.92}
\footnotesize
\begin{tabular}{@{}lcccccccccccccc@{}}
\toprule
& \multicolumn{7}{c}{\textbf{Suite A: Held-Out GeoDash}} & \multicolumn{7}{c}{\textbf{Suite B: Held-Out Doodle Jump}} \\
\cmidrule(lr){2-8}\cmidrule(lr){9-15}
Ablation Variant & \multicolumn{2}{c}{Task 1} & \multicolumn{2}{c}{Task 2} & \multicolumn{2}{c}{\textbf{Mean}} & Resets$\downarrow$ & \multicolumn{2}{c}{Task 1} & \multicolumn{2}{c}{Task 2} & \multicolumn{2}{c}{\textbf{Mean}} & Resets$\downarrow$ \\
& SR & PG & SR & PG & SR & PG & & SR & PG & SR & PG & SR & PG & \\
\midrule
Token SFT (Offline Token BC) & 0/30 & 65.1 & 0/30 & 31.1 & 0.0 & 48.1 & 462 & 5/30 & 65.5 & 0/30 & 15.9 & 8.3 & 40.7 & 85 \\
+ Complete-Action Alignment & 11/30 & 85.6 & 0/30 & 53.7 & 18.3 & 69.7 & 57 & 2/30 & 65.9 & 0/30 & 27.8 & 3.3 & 46.8 & 71 \\
+ Selective MOPD (Uniform) & 7/30 & 79.4 & 0/30 & 47.6 & 11.7 & 63.5 & 87 & 8/30 & 57.7 & 0/30 & \textbf{41.9} & 13.3 & 49.8 & 41 \\
\rowcolor{cvprblue!10}
\textbf{+ Conflict-Aware Consolidation (Full \method)} & \textbf{26/30} & \textbf{95.5} & 0/30 & \textbf{58.0} & \textbf{43.3} & \textbf{76.7} & \textbf{36} & \textbf{24/30} & \textbf{91.9} & 0/30 & 39.4 & \textbf{40.0} & \textbf{65.7} & \textbf{12} \\
\bottomrule
\end{tabular}
\vspace{-0.5em}
\end{table*}

\paragraph{Performance on Suite~A (temporal control).}
On \emph{GeoDash}, zero-shot Base VLM triggers premature jumps, crashing into
spikes at $t \approx 68$ and resetting to $0\%$ in \cref{fig:results}(a4) (0/30
Task-1 SR), while prompted VLMs (Pause-and-Plan CoT at 13/30 SR, 73.8 Mean PG;
Few-shot ICL at 5/30 SR) remain sensitive to timing drift. Single-source
specialists expose the underlying teacher conflict: \emph{Boxel Rebound}
transfers reactive obstacle-jumping timing (14/30 SR, 67.0 Mean PG), whereas
\emph{Stack} reflects block-alignment dynamics that fail to transfer (0/30 SR)---as failed attempts respawn right before the first spike and crash within two decisions, mis-timed jumps incur ${\sim}3{,}000$ resets across the 6,000 total steps of Tasks~1--2.
Because Multi-Task PCGrad projects conflicting gradients onto normal planes with
equal task priority, updates from \emph{Stack} attenuate transferable timing
signals (0/30 SR), while parameter mergers (peaking at 19/30 SR for TIES-Merging
and 73.3 Mean PG for DELLA) and Feature-Routing MoE (17/30 SR) still suffer from
adapter interference. In contrast, by resolving teacher conflict in aligned
action space rather than blending divergent weights, \method waits on flat
segments and jumps cleanly over hazards (\cref{fig:results}(a1--a3)), reaching
the Task-1 threshold in \cref{fig:results}(a4) with \textbf{26/30} Task-1 SR,
\textbf{76.7} Mean PG, and the fewest resets (\textbf{36})---outperforming the
strongest baselines in both success rate (19/30 for TIES-Merging) and progress
(73.8 Mean PG for CoT).

\paragraph{Performance on Suite~B (directional steering).}
Suite~B requires sustained lateral alignment with ascending platforms on
\emph{Doodle Jump} (\cref{fig:results}(b1--b4)). Prompted VLMs reach at most
12/30 Task-1 SR (Few-shot ICL, 48.1 Mean PG), while among source specialists,
\emph{NS-Shaft} shares compatible platform-landing dynamics (6/30 SR) whereas
\emph{Breakout} and \emph{Cubefield} fail to transfer (0/30 SR). Interpolating
LoRA deltas in parameter space (Task Arithmetic, TIES-Merging, DARE-TIES, DELLA,
RegMean) implicitly assumes linear connectivity~\cite{fu2026mergevla}; combining
\emph{NS-Shaft} with opposing steering adapters cancels lateral control, leaving
policies bouncing in place on the starting platform (0\% PG and thus 0 fall
resets), while Multi-Task PCGrad (0/30 SR) and test-time Feature-Routing MoE (4/30 SR under
visual shift) likewise struggle. In \cref{fig:results}(b4), Base VLM stalls near
the bottom and Action-aligned SFT falls at $t = 50$ after drifting off-course;
in contrast, \method suppresses conflicting corrections in action space to
correct lateral drift (\cref{fig:results}(b1--b3)) and reach $100\%$ progress
(\cref{fig:results}(b4)), attaining \textbf{24/30} Task-1 SR (\textbf{65.7} Mean
PG, \textbf{12} resets) and doubling the strongest baseline SR
(Few-shot ICL, 12/30).

\subsection{Ablation Studies}
\label{sec:ablations}

\paragraph{Effect of complete-action alignment.}
\Cref{tab:component-ablation} evaluates each component from offline Token SFT,
which exhibits marked closed-loop instability (462 and 85 resets in Suites~A
and~B) as subword cross-entropy biases predictions toward shorter or lexically
favored strings. Normalizing command completion likelihoods over each game's
legal action set $\mathcal{A}_g$ via \emph{Complete-Action Alignment} mitigates
subword length bias and forms a calibrated categorical policy, cutting Suite~A
resets from 462 to 57 (Task-1 SR $0/30{\to}11/30$) and lifting Suite~B Mean PG
from 40.7 to 46.8, though offline alignment remains prone to rollout drift
(2/30 Suite-B SR).

\begin{table}[!b]
\vspace{-0.5em}
\centering
\caption{\textbf{Cross-game disagreement $C_{g,g'}$ and $\gamma_g$ weights.}
Lower disagreement $\bar{C}_g$ yields higher weight $\gamma_g$.}
\label{tab:supp-gate-matrix}
\setlength{\tabcolsep}{2.52pt}
\renewcommand{\arraystretch}{0.90}
\footnotesize
\begin{tabular}{@{}lccccccc@{}}
\toprule
Specialist $\pi_{T_g}$ & \multicolumn{3}{c}{Val.\ NLL $C_{g, g'}\downarrow$} & $\bar{C}_g\downarrow$ & $z_g$ & $\rho_g\uparrow$ & $\gamma_g\uparrow$ \\
\cmidrule(lr){2-4}
\textit{Suite A} & \emph{Boxel} & \emph{Stack} & -- & & & & \\
\rowcolor{cvprblue!10}
\textbf{Boxel Rebound} & 0.059 & 0.706 & -- & \textbf{0.383} & $+1.00$ & \textbf{0.843} & \textbf{1.69} \\
Stack & 0.806 & 0.468 & -- & 0.637 & $-1.00$ & 0.157 & 0.31 \\
\midrule
\textit{Suite B} & \emph{Breakout} & \emph{Cube.} & \emph{NS-Shaft} & & & & \\
\rowcolor{cvprblue!10}
\textbf{NS-Shaft} & 2.610 & 1.725 & 0.050 & \textbf{1.462} & $+1.18$ & \textbf{0.665} & \textbf{2.00} \\
Breakout & 3.050 & 0.986 & 0.984 & 1.673 & $+0.09$ & 0.247 & 0.74 \\
Cubefield & 3.155 & 0.562 & 2.098 & 1.938 & $-1.27$ & 0.088 & 0.26 \\
\bottomrule
\end{tabular}
\end{table}

\begin{table*}[!t]
\centering
\caption{\textbf{Causal visual-feedback interventions and open-loop controls.}
Replacing live frames $o_{t-h+1:t}$ with a frozen initial frame $o_0$ or
vision-free rules markedly degrades closed-loop success.}
\label{tab:causal-counterfactuals}
\setlength{\tabcolsep}{2.40pt}
\renewcommand{\arraystretch}{0.88}
\footnotesize
\begin{tabular}{@{}llcccccccccccccc@{}}
\toprule
& & \multicolumn{7}{c}{\textbf{Suite A: Held-Out GeoDash}} & \multicolumn{7}{c}{\textbf{Suite B: Held-Out Doodle Jump}} \\
\cmidrule(lr){3-9}\cmidrule(lr){10-16}
Policy Variant & Visual Input \& Decision Mode & \multicolumn{2}{c}{Task 1} & \multicolumn{2}{c}{Task 2} & \multicolumn{2}{c}{\textbf{Mean}} & Resets$\downarrow$ & \multicolumn{2}{c}{Task 1} & \multicolumn{2}{c}{Task 2} & \multicolumn{2}{c}{\textbf{Mean}} & Resets$\downarrow$ \\
& & SR & PG & SR & PG & SR & PG & & SR & PG & SR & PG & SR & PG & \\
\midrule
Passive \texttt{wait} & None (constant \texttt{wait}) & 0/30 & 30.0 & 0/30 & 15.0 & 0.0 & 22.5 & 1952 & 0/30 & 0.0 & 0/30 & 0.0 & 0.0 & 0.0 & 0 \\
Periodic cadence & None (fixed \texttt{jump} / left--right cycle) & 16/30 & 85.9 & 0/30 & \textbf{65.6} & 26.7 & 75.7 & 313 & 0/30 & 35.8 & 0/30 & 18.7 & 0.0 & 27.2 & 14 \\
\method (frozen $o_0$) & Frozen initial frame $o_0$ (static) & 1/30 & 92.3 & 0/30 & 46.0 & 1.7 & 69.1 & 69 & 0/30 & 26.8 & 0/30 & 2.1 & 0.0 & 14.4 & 45 \\
\midrule
\rowcolor{cvprblue!10}
\textbf{\method (live)} & \textbf{Live frames $o_{t-h+1:t}$ (closed-loop)} & \textbf{26/30} & \textbf{95.5} & 0/30 & 58.0 & \textbf{43.3} & \textbf{76.7} & \textbf{36} & \textbf{24/30} & \textbf{91.9} & 0/30 & \textbf{39.4} & \textbf{40.0} & \textbf{65.7} & \textbf{12} \\
\bottomrule
\end{tabular}
\vspace{-0.5em}
\end{table*}

\paragraph{Effect of selective on-policy retention.}
Adding \emph{Selective MOPD (Uniform)} uses these calibrated uncertainty margins
to retain teacher corrections $\mathcal{Q}$ on student-visited error and
low-margin states, consolidating $\mathcal{D}_0 \cup \mathcal{Q}$ under \cref{eq:amopd} with equal teacher weights ($\gamma_g{=}1$). Supervising such boundary states mitigates rollout drift in
Suite~B, lifting Task-1 SR from 2/30 to 8/30 and cutting resets ($71{\to}41$).
However, uniform pooling over $\mathcal{Q}$
exposes \emph{multi-teacher policy conflict}: in Suite~A, incompatible
corrections from \emph{Stack} interfere with \emph{Boxel Rebound}'s jump timing
and drop Task-1 SR from 11/30 to 7/30, while in Suite~B opposing steering
signals from \emph{Breakout} and \emph{Cubefield} lower Task-1 PG from 65.9 to
57.7, motivating conflict-aware reweighting over $\mathcal{Q}$.

\paragraph{Effect of conflict-aware multi-teacher consolidation.}
Incorporating \emph{Conflict-Aware Consolidation} (\textbf{Full \method}) resolves conflict
in $\mathcal{D}_0 \cup \mathcal{Q}$ without target data by cross-evaluating the
aligned specialists to upweight compatible teachers
and suppress conflicting corrections, lifting Task-1 SR from $7/30{\to}\mathbf{26/30}$
($63.5{\to}\mathbf{76.7}$ Mean PG; resets $87{\to}\mathbf{36}$) in Suite~A and
$8/30{\to}\mathbf{24/30}$ ($49.8{\to}\mathbf{65.7}$ Mean PG; resets
$41{\to}\mathbf{12}$) in Suite~B. \emph{Selective MOPD}'s higher Suite~B Task-2
PG reflects peak-bounce variance across 41 resets (vs.\ 12) rather than sustained climbing.

%% file: sections/05_discussion.tex
\section{Further Analysis and Generalization}
\label{sec:analysis}

\subsection{Multi-Teacher Policy Conflict Analysis}
\label{sec:gate-analysis}

\paragraph{Asymmetric inter-teacher disagreement.}
\Cref{tab:supp-gate-matrix} inspects $C_{g,g'}$ and $\gamma_g$
(\cref{eq:cross-source-nll,eq:compatibility-gate}). In Suite~A, \emph{Boxel
Rebound} fits its own states far more reliably than \emph{Stack} ($0.059$ vs.\
$0.468$ diagonal NLL) and incurs lower cross-game loss ($0.706$ vs.\ $0.806$),
yielding lower $\bar{C}_g$ ($0.383$ vs.\ $0.637$) as in-game reliability and
peer compatibility both favor \emph{Boxel Rebound}. In Suite~B, \emph{Breakout}
includes additional rare long-stroke moves (\texttt{*\_long};
\cref{tab:supp-teachers}), and predicting wall-bounced ball paths from two
frames is visually ambiguous. Consequently, $\pi_{T_{\text{Breakout}}}$ assigns
near-zero probability to rare \texttt{*\_long} moves on the action-balanced
validation split ($3.050$ NLL vs.\ $2.610$ for $\pi_{T_{\text{NS-Shaft}}}$,
which does not suppress unseen \texttt{*\_long} actions), whereas
\emph{NS-Shaft} fits its own three-action control cleanly ($0.050$). Meanwhile,
\emph{Cubefield} fits its own dodging ($0.562$) yet incurs $2.098$ on
\emph{NS-Shaft} from opposing rules (dodging vs.\ platform landing).

\paragraph{Target-free teacher identification.}
Computed solely from source validation splits, the inferred weights $\gamma_g$
accurately reflect standalone zero-shot transfer (\cref{tab:main-summary}): in
Suite~A, \emph{Boxel Rebound} is upweighted over \emph{Stack} ($\gamma_g{=}1.69$
vs.\ $0.31$, corresponding to $67.0$ vs.\ $22.5$ Mean PG on \emph{GeoDash}); in
Suite~B, \emph{NS-Shaft} ($\gamma_g{=}2.00$) is prioritized over
\emph{Breakout} ($0.74$) and \emph{Cubefield} ($0.26$), mirroring their
standalone Mean PG of $44.1$, $7.1$, and $0.0$ on \emph{Doodle Jump}. Moreover, multi-teacher consolidation markedly outperforms the
strongest standalone specialist in each suite (\textbf{26/30} vs.\ 14/30 Task-1
SR in Suite~A; \textbf{24/30} vs.\ 6/30 in Suite~B): retaining $\mathcal{D}_0$
with prior floor $\epsilon{=}0.1$ preserves multi-game visual diversity while
selectively suppressing conflicting boundary updates.

\begin{figure}[!b]
    \vspace{-0.8em}
    \centering
    \includegraphics[width=0.94\linewidth]{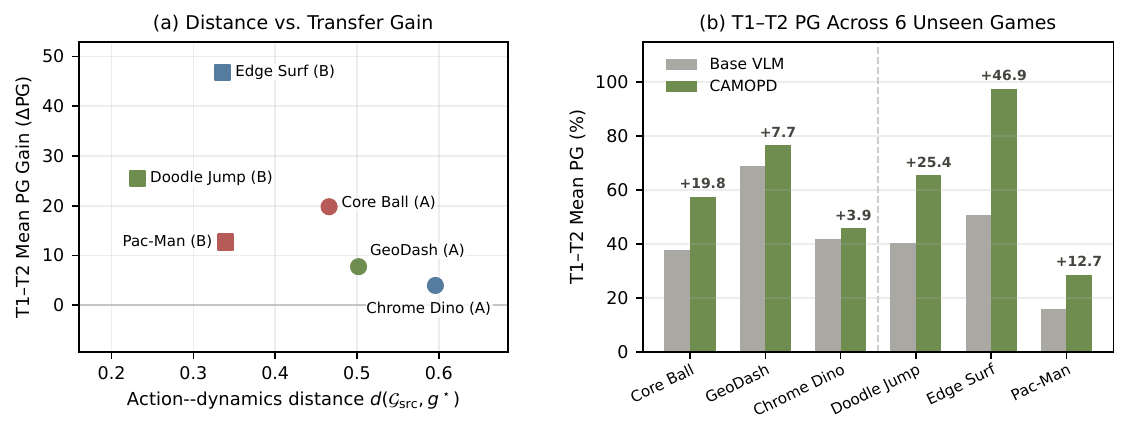}
    \vspace{-0.8em}
    \caption{\textbf{Cross-genre transfer across six unseen target games.}
    (a) Action--dynamics distance $d(\sourcegames, g^\star)$ vs.\ progress gain
    $\Delta\mathrm{PG}$. (b) Mean PG of Base VLM and \method ordered by
    distance per suite.}
    \label{fig:genre-distance}
\end{figure}

\vspace{-0.3em}
\subsection{Causal Visual Interventions}
\label{sec:causal-results}
\vspace{-0.2em}

\Cref{tab:causal-counterfactuals} tests whether transfer relies on live visual
feedback rather than open-loop priors. Vision-free controls fail to sustain
play: constant \texttt{wait} stalls, while a fixed periodic cycle inflates
Suite~A Task-2 PG only via 313 crash resets ($8.7\times$ \method), dropping
Task-1 SR to 16/30 in Suite~A and 0/30 in Suite~B. Crucially, freezing input at
the initial frame $o_0$ collapses Task-1 SR from \textbf{26/30 to 1/30} in
Suite~A and from \textbf{24/30 to 0/30} in Suite~B. These results confirm that
live visual feedback is essential for closed-loop timing and drift correction,
and show that \method grounds decisions in visual dynamics rather than action
rhythms.

\vspace{-0.3em}
\subsection{Cross-Genre Distance and Transfer Boundaries}
\label{sec:teacher-genre-analysis}
\vspace{-0.2em}

\Cref{fig:genre-distance} examines cross-genre transfer across six unseen
target games as a function of the action--dynamics distance
$d(\sourcegames, g^\star)=\frac12(d_{\mathrm{act}}+d_{\mathrm{dyn}})$
(Appendix~\ref{sec:supp-genre}). Overall, transfer gains tend to diminish as
$d(\sourcegames, g^\star)$ increases. Within Suite~A, progress gain decreases
monotonically with distance from \emph{Core Ball} ($d{=}0.466$, $+19.8$ PG) to
\emph{GeoDash} ($d{=}0.502$, $+7.7$ PG) and \emph{Chrome Dino} ($d{=}0.596$,
$+3.9$ PG). In Suite~B, gains similarly decline from \emph{Doodle Jump}
($d{=}0.232$, $+25.4$ PG) to \emph{Pac-Man} ($d{=}0.339$, $+12.7$ PG).
Notably, \emph{Edge Surf} ($d{=}0.336$) achieves a larger $+46.9$ PG gain, as
its obstacle-avoidance dynamics closely match the consolidated lateral steering
prior.

\vspace{-0.3em}
\section{Conclusion}
\label{sec:conclusion}
\vspace{-0.2em}

We present \method, a conflict-aware multi-teacher on-policy distillation
framework that consolidates heterogeneous source teachers into a single
pixel-only VLM without target data. Built on length-normalized complete-action
distributions, \method retains boundary corrections via calibrated margins and
cross-evaluates visual specialist probes across peer source games to upweight
compatible teachers and suppress conflicting updates. On GameWorld's temporal
and directional suites, \method reaches 86.7\% and 80.0\% Task-1 zero-shot
success, surpassing prompting, single-teacher, merging, and multi-task methods.

\vspace{-0.3em}
\section{Limitations}
\vspace{-0.2em}
First, following GameWorld~\cite{ouyang2026gameworld}, evaluation pauses
execution during inference over discrete commands rather
than real-time continuous streams. Second, the student
conditions on a short $h$-frame window ($h \in \{1, 2\}$) without explicit
long-horizon episodic memory. Extending \method to streaming control and
memory-augmented policies remains future work; further exploring
state-adaptive compatibility weighting beyond source-level $\gamma_g$ is also a
promising direction.

%% file: sections/07_llm_usage.tex
\section*{LLM Usage Statement}
OpenAI Codex was used to assist with code generation and completion, OpenAI GPT
models with related-literature discovery, and Google Gemini with manuscript
drafting and editing. The authors led the research process, problem
formulation, computational experiments, and factual verification, and take full
responsibility for all scientific claims, results, references, and the final
manuscript.

%% file: sections/06_supplement.tex
\section{Algorithm Pseudocode}
\label{sec:supp-alg}
\vspace{0.1em}
Algorithm~1 summarizes the end-to-end training and zero-shot deployment pipeline of \method (\cref{sec:method}).
\vspace{0.4em}

\begin{center}
\small
\setlength{\tabcolsep}{5pt}
\renewcommand{\arraystretch}{1.14}
\begin{tabular}{@{}r@{\hspace{8pt}}p{0.94\textwidth}@{}}
\toprule
\multicolumn{2}{@{}l@{}}{\textbf{Algorithm 1:} Conflict-Aware Multi-Teacher On-Policy Distillation in Aligned Action Spaces (\method)} \\
\midrule
\multicolumn{2}{@{}p{\textwidth}@{}}{\textbf{Input:} Source games $\sourcegames = \{g_1, \dots, g_M\}$ with rule teachers $\{\tau_g\}_{g \in \sourcegames}$ and legal action sets $\{\mathcal{A}_g\}_{g \in \sourcegames}$; base VLM with shared LoRA parameters $\theta$; retention thresholds $(c_{\min}, \delta)$; conflict-aware reweighting parameters $(w_q, c_{\mathrm{floor}}, \beta, \epsilon)$; consolidation loss parameters $(\lambda, T_{\mathrm{ref}})$.} \\
\multicolumn{2}{@{}p{\textwidth}@{}}{\textbf{Output:} Consolidated pixel-only student policy $\pi_\theta$ for zero-shot deployment on unseen target games $g^\star \notin \sourcegames$.} \\
\midrule
\multicolumn{2}{@{}l@{}}{\textbf{Complete-Action Alignment (\cref{sec:action-alignment})}} \\
1: & Roll out each teacher $\tau_g(s_t) \to (a_t^\star, c_t)$ on $g \in \sourcegames$ to collect offline bank $\mathcal{D}_0 = \bigcup_{g \in \sourcegames} \mathcal{D}_g$ and action-balanced validation splits $\{\mathcal{D}_g^{\mathrm{val}}\}_{g \in \sourcegames}$. \\
2: & Normalize completion likelihoods over $\mathcal{A}_g$: $s_\theta(a \mid x) = \frac{1}{L_a} \sum_{\ell=1}^{L_a} \log p_\theta(y_\ell(a) \mid x, y_{<\ell}(a))$ \hfill $\triangleright$ \cref{eq:score} \\
3: & Define calibrated categorical policy $\pi_\theta^{(T)}(a \mid x, \mathcal{A}_g) = \exp(s_\theta(a \mid x)/T) \big/ \sum_{a' \in \mathcal{A}_g} \exp(s_\theta(a' \mid x)/T)$ \hfill $\triangleright$ \cref{eq:policy} \\
4: & Train initial shared student $\theta_{\mathrm{SFT}}$ on $\mathcal{D}_0$ and frozen visual specialists $\{\pi_{T_g}\}_{g \in \sourcegames}$ on each $\mathcal{D}_g$ by minimizing $\mathcal{L}_{\mathrm{SFT}}(\theta)$ \hfill $\triangleright$ \cref{eq:sft} \\
\addlinespace[2pt]
\multicolumn{2}{@{}l@{}}{\textbf{Selective On-Policy Retention (\cref{sec:mopd})}} \\
5: & Initialize student $\theta \leftarrow \theta_{\mathrm{SFT}}$ and on-policy correction buffers $\mathcal{Q}_g \leftarrow \varnothing$ for all $g \in \sourcegames$. \\
6: & \textbf{for} each source game $g \in \sourcegames$ and closed-loop rollout step $t = 1, \dots, H$ \textbf{do} \\
7: & \quad Execute $\hat{a}_t = \arg\max_{a \in \mathcal{A}_g} \pi_\theta(a \mid x_t, \mathcal{A}_g)$ with calibrated uncertainty margin $m_\theta(x_t) = s_\theta(\hat{a}_t \mid x_t) - \max_{a \ne \hat{a}_t} s_\theta(a \mid x_t)$. \\
8: & \quad Query privileged rule teacher $(a_t^\star, c_t) \leftarrow \tau_g(s_t)$ on underlying simulator state $s_t$. \\
9: & \quad \textbf{if} $c_t \ge c_{\min}$ \textbf{and} $(\hat{a}_t \ne a_t^\star \;\lor\; m_\theta(x_t) < \delta)$ \textbf{then} $\mathcal{Q}_g \leftarrow \mathcal{Q}_g \cup \{(x_t, \mathcal{A}_g, a_t^\star, c_t)\}$ \hfill $\triangleright$ Selective retention (\cref{eq:query}) \\
10: & \textbf{end for} \\
11: & Form hybrid replay bank $\mathcal{D}_0 \cup \mathcal{Q}$ with $\mathcal{Q} = \bigcup_{g \in \sourcegames} \mathcal{Q}_g$ and on-policy indicator $q_i = \mathbb{I}[i \in \mathcal{Q}]$. \\
\addlinespace[2pt]
\multicolumn{2}{@{}l@{}}{\textbf{Conflict-Aware Multi-Teacher Consolidation (\cref{sec:conflict-consolidation})}} \\
12: & Cross-evaluate each specialist $\pi_{T_g}$ across $\mathcal{D}_{g'}^{\mathrm{val}}$ ($g' \in \sourcegames$) to obtain disagreement matrix $C_{g,g'}$ and mean $\bar{C}_g = \frac{1}{M}\sum_{g'} C_{g,g'}$ \hfill $\triangleright$ \cref{eq:cross-source-nll} \\
13: & Compute $z_g = -\beta(\bar{C}_g - \mu_C)/\sigma_C$, Gibbs weights $\rho_g = (1-\epsilon)\operatorname{softmax}(z)_g + \epsilon/M$, and teacher multipliers $\gamma_g = M\rho_g$ \hfill $\triangleright$ \cref{eq:compatibility-gate} \\
14: & With $\tilde{c}_i = \max(c_{\mathrm{floor}}, c_i)$, set $u_i^{\mathrm{ca}} = \big(1 + (w_q\gamma_{g_i} - 1)q_i\big)\tilde{c}_i$ and normalize $P_{\mathrm{global}}(i) = u_i^{\mathrm{ca}} \big/ \sum_{j \in \mathcal{D}_0 \cup \mathcal{Q}} u_j^{\mathrm{ca}}$ \hfill $\triangleright$ \cref{eq:grad-decomp} \\
15: & Freeze $\theta_{\mathrm{ref}} \leftarrow \theta_{\mathrm{SFT}}$ and update $\theta$ on minibatches $i \sim P_{\mathrm{global}}$ by minimizing $\mathcal{L}(\theta) = \mathcal{L}_{\mathrm{action}}(\theta) + \lambda\,\mathcal{L}_{\mathrm{anchor}}(\theta)$ \hfill $\triangleright$ \cref{eq:amopd} \\
\addlinespace[2pt]
\multicolumn{2}{@{}l@{}}{\textbf{Zero-Shot Target Deployment}} \\
16: & Discard $\{\tau_g\}$ and $\{\pi_{T_g}\}$; on unseen game $g^\star \notin \sourcegames$, execute $\hat{a}_t = \arg\max_{a \in \mathcal{A}_{g^\star}} \pi_\theta(a \mid x_t, \mathcal{A}_{g^\star})$ from live screenshots $o_{t-h+1:t}$. \\
\bottomrule
\end{tabular}
\end{center}
\vspace{0.5em}

\section{Action--Dynamics Domain Distance Formulation}
\label{sec:supp-genre}

To characterize cross-genre transfer boundaries in \cref{sec:teacher-genre-analysis} and \cref{fig:genre-distance}, we collect short uniform-random rollouts $\{(o_t, a_t, o'_t)\}$ separate from policy training and extract $\ell_2$-normalized visual embeddings $\phi(o) \in \mathbb{R}^d$ using the frozen Qwen3.5-4B vision tower. For any pair of games $(g_i, g_j)$, we compute two complementary distance axes:
\begin{enumerate}
    \item \textbf{Action-semantic distance ($d_{\mathrm{act}}$):} mapping surface commands in $\mathcal{A}_g$ to coarse control roles (\texttt{wait}, vertical/event triggers, lateral/vertical turns), we take an equal-weight combination of role-set Jaccard distance between $\mathcal{A}_{g_i}$ and $\mathcal{A}_{g_j}$ and the normalized cosine distance between role-conditioned feature displacements $\mathbb{E}[\phi(o'_t) - \phi(o_t) \mid e]$ matched across shared roles $e$ ($1.0$ if disjoint).
    \item \textbf{Transition dynamics distance ($d_{\mathrm{dyn}}$):} squared maximum mean discrepancy ($\mathrm{MMD}^2$) with a Gaussian radial basis function (RBF) kernel over frame-to-frame transition vectors $\Delta\phi_t = \phi(o'_t) - \phi(o_t)$, capturing temporal motion similarity.
\end{enumerate}
Normalizing each axis to $[0, 1]$ across evaluated game pairs yields the pairwise and suite-to-target action--dynamics distance:
\begin{equation}
\begin{split}
d(g_i,g_j)&=\tfrac12
\big(d_{\mathrm{act}}(g_i,g_j)+d_{\mathrm{dyn}}(g_i,g_j)\big),\\
d(\sourcegames,g^\star)&=\frac{1}{|\sourcegames|}
\sum_{g\in\sourcegames}d(g,g^\star).
\end{split}
\label{eq:distance}
\end{equation}
Intuitively, \cref{eq:distance} focuses on functional control mechanics rather than static appearance: pretrained VLM encoders already generalize across background colors and sprite textures, whereas zero-shot policy transfer hinges on whether actions induce comparable visual displacements ($d_{\mathrm{act}}$) under similar frame-to-frame motion pacing ($d_{\mathrm{dyn}}$). Small values on both axes indicate that consolidated timing or steering priors act on aligned transition dynamics, whereas additional target control roles (larger $d_{\mathrm{act}}$) or shifted motion distributions (larger $d_{\mathrm{dyn}}$) narrow the transferable skill subspace, and averaging over $g \in \sourcegames$ summarizes proximity to the multi-source pool. Note that $d(\sourcegames, g^\star)$ is computed only for post-hoc analysis of unseen target games and is not used during student training or conflict-aware multi-teacher consolidation. \Cref{tab:supp-genre-distance} reports the native action sets, per-axis distances, and zero-shot T1--T2 Mean PG across all six unseen target games; during cross-genre evaluation, each suite scores the target game's native temporal (\emph{Core Ball}: \texttt{\{wait, shoot\}}, \emph{Chrome Dino}: \texttt{\{wait, jump\}}) or lateral (\emph{Edge Surf}: \texttt{\{wait, turn\_left/right\}}, \emph{Pac-Man}: \texttt{\{wait, move\_left/right\}}) command subset matching the source control axis.

\begin{table}[!ht]
\centering
\caption{\textbf{Action--dynamics distances and zero-shot transfer across six unseen target games.}
Games are ordered by $d(\sourcegames, g^\star)$ within each suite, reporting native game action spaces $\mathcal{A}_{g^\star}$ for $d_{\mathrm{act}}$, per-axis distances, and T1--T2 Mean PG (\%) under each suite's transferred temporal (Suite~A) or lateral (Suite~B) control actions.}
\label{tab:supp-genre-distance}
\setlength{\tabcolsep}{0pt}
\renewcommand{\arraystretch}{0.96}
\footnotesize
\begin{tabular*}{\textwidth}{@{\extracolsep{\fill}}lllcccccc@{}}
\toprule
Suite & Target Game ($g^\star$) & Native Actions ($\mathcal{A}_{g^\star}$) & $d_{\mathrm{act}}$ & $d_{\mathrm{dyn}}$ & $d(\sourcegames, g^\star)$ & Base PG & \method PG & $\Delta\mathrm{PG}$ \\
\midrule
\multirow{3}{*}{Suite A}
& \textbf{Core Ball} & \texttt{\{wait, shoot\}} & 0.610 & 0.321 & 0.466 & 37.8 & \textbf{57.6} & \textbf{+19.8} \\
& \textbf{GeoDash} & \texttt{\{wait, jump\}} & 0.576 & 0.427 & 0.502 & 69.0 & \textbf{76.7} & \textbf{+7.7} \\
& \textbf{Chrome Dino} & \texttt{\{wait, jump, duck, start\}} & 0.718 & 0.474 & 0.596 & 42.1 & \textbf{46.0} & \textbf{+3.9} \\
\midrule
\multirow{3}{*}{Suite B}
& \textbf{Doodle Jump} & \texttt{\{wait, move\_left/right\}} & 0.415 & 0.049 & 0.232 & 40.3 & \textbf{65.7} & \textbf{+25.4} \\
& \textbf{Edge Surf} & \texttt{\{wait, turn\_left/right, boost, slow\_down\}} & 0.620 & 0.051 & 0.336 & 50.7 & \textbf{97.6} & \textbf{+46.9} \\
& \textbf{Pac-Man} & \texttt{\{wait, move\_left/right/up/down\}} & 0.627 & 0.051 & 0.339 & 16.0 & \textbf{28.7} & \textbf{+12.7} \\
\bottomrule
\end{tabular*}
\end{table}

\section{Prompt Template and Privileged Source Teachers}
\label{sec:supp-prompts-teachers}

\paragraph{Prompt template and complete-action completion scoring.}
In Complete-Action Alignment (\cref{sec:action-alignment}), each multimodal context $x_t = (o_{t-h+1:t}, q_g)$ is formatted under Qwen3.5-4B's chat template with reasoning disabled (\texttt{enable\_thinking=False}) so that inference and training operate on executable semantic commands. \Cref{tab:supp-prompt-template} illustrates the prompt structure and how Complete-Action Alignment (\cref{eq:score}) scores the legal command completion tokens $y(a)$ while masking prompt tokens.

\begin{table}[!ht]
\vspace{-0.3em}
\centering
\caption{\textbf{Multimodal prompt template and complete-action completion scoring.}
Token-level SFT biases scoring across variable-length subword sequences, whereas Complete-Action Alignment masks the prompt and normalizes log-probabilities over the $L_a$ action completion tokens.}
\label{tab:supp-prompt-template}
\setlength{\tabcolsep}{4pt}
\renewcommand{\arraystretch}{1.02}
\footnotesize
\begin{tabular}{@{}p{\textwidth}@{}}
\toprule
\textbf{System Prompt ($q_g$):} \\
\texttt{You are controlling the GameWorld game $g$. Task: <task\_prompt> Choose exactly one legal action from: $\mathcal{A}_g$. Return exactly one JSON object and no other text:\allowbreak{} \{"tool\_name":"<action>",\allowbreak"arguments":\{\}\}.} \\
\midrule
\textbf{User Input ($o_{t-h+1:t}$):} $\langle\texttt{image}_{t-h+1}\rangle \dots \langle\texttt{image}_t\rangle$ \\
\midrule
\textbf{Candidate Action Completion Suffixes $y(a)$ for $a \in \mathcal{A}_g$:} \\
$\bullet$ $a = \texttt{wait}$: \;\; \texttt{\{"tool\_name":"wait",\allowbreak"arguments":\{\}\}} \\
$\bullet$ $a = \texttt{move\_right}$: \;\; \texttt{\{"tool\_name":"move\_right",\allowbreak"arguments":\{\}\}} \\
\textbf{Scoring (\cref{eq:score}):} Mask all system/user prompt tokens and compute $s_\theta(a \mid x) = \frac{1}{L_a}\sum_{\ell=1}^{L_a} \log p_\theta(y_\ell(a) \mid x, y_{<\ell}(a))$. \\
\bottomrule
\end{tabular}
\vspace{-0.4em}
\end{table}

\paragraph{Privileged source rule teachers.}
During source data collection only ($g \in \sourcegames$), each source game is paired with a programmatic rule teacher $\tau_g(s_t) \to (a_t^\star, c_t)$ that reads simulator state variables $s_t \in \mathcal{S}_g$ and outputs reference action $a_t^\star \in \mathcal{A}_g$ along with criticality confidence $c_t \in (0, 1]$ ($c_{\min} = 0.75$). \Cref{tab:supp-teachers} summarizes the privileged state inputs, decision rules, and confidence assignments of all five source teachers across Suites~A and~B.

\begin{table}[!ht]
\centering
\caption{\textbf{Summary of privileged programmatic rule teachers $\tau_g(s_t)$ across source games ($g \in \sourcegames$).}
Privileged simulator states $s_t$ are queried only during source data collection (incomplete or unreachable-platform simulator frames emit $c_t \in [0.50, 0.55]$ and are filtered out during retention); at deployment, all teachers are discarded.}
\label{tab:supp-teachers}
\setlength{\tabcolsep}{0pt}
\renewcommand{\arraystretch}{0.96}
\footnotesize
\begin{tabular*}{\textwidth}{@{\extracolsep{\fill}}llp{0.222\textwidth}lp{0.335\textwidth}@{}}
\toprule
Suite & Source Game ($g$) & Legal Action Set ($\mathcal{A}_g$) & Privileged State ($s_t$) & Rule Teacher Decision \& Confidence Logic $\tau_g(s_t) \to (a_t^\star, c_t)$ \\
\midrule
\multirow{2}{*}{Suite A}
& \textbf{Boxel Rebound} & \texttt{\{wait, jump\}} & Ground flag, tile grid, subcell $x$ & Scans a short lookahead window ($[2, 5]$ tiles) for spikes, gaps, or walls while grounded: outputs \texttt{jump} ($c_t \in [0.80, 0.96]$) when a hazard enters the jump window, and \texttt{wait} ($c_t \in [0.90, 0.98]$) when airborne or on clear ground. \\
& \textbf{Stack} & \texttt{\{wait, drop\_block\}} & Moving \& top block coords/sizes & Computes horizontal overlap ratio with the top stack block: outputs \texttt{drop\_block} ($c_t \in [0.75, 1.00]$) when overlap ratio $\ge 0.90$, and \texttt{wait} ($c_t \in [0.82, 0.92]$) while approaching alignment. \\
\midrule
\multirow{3}{*}{Suite B}
& \textbf{Breakout} & \texttt{\{wait[\_long],} \texttt{move\_left/right[\_long]\}} & Ball $(x,y,v_x,v_y)$, paddle $x$ & Reflects the descending ball trajectory against side walls to predict the paddle-plane intercept: emits \texttt{move\_left/right} (or \texttt{*\_long} on large intercept gaps, $2.5\%$ of frames; $c_t \in [0.80, 0.99]$) when displacement exceeds deadzone, else \texttt{wait} ($c_t{=}0.97$; runtime action \texttt{wait\_long} is unused by $\tau_g$). \\
& \textbf{NS-Shaft} & \texttt{\{wait, move\_left/right\}} & Player $(x,y,v_y)$, platform list & Selects the highest reachable safe platform below the player: steers horizontally toward the platform center ($c_t \in [0.80, 0.99]$) and outputs \texttt{wait} ($c_t{=}0.96$) once aligned within the landing margin. \\
& \textbf{Cubefield} & \texttt{\{wait, move\_left/right\}} & Corridor offset, obstacle cones & Projects forward collision cones against approaching cubes: steers toward the wider clearance gap ($c_t \in [0.84, 0.99]$) under imminent collision risk, else recenters or emits \texttt{wait} ($c_t \in [0.86, 0.95]$). \\
\bottomrule
\end{tabular*}
\vspace{-0.4em}
\end{table}